\documentclass[10pt,twocolumn,letterpaper]{article}

\usepackage{cvpr}              

\usepackage{graphicx}
\usepackage{amsmath}
\usepackage{amssymb}
\usepackage{booktabs}
\usepackage{verbatim}
\usepackage{amsmath}
\usepackage{array}
\usepackage{multirow}

\usepackage[pagebackref,breaklinks,colorlinks]{hyperref}

\usepackage[capitalize]{cleveref}
\crefname{section}{Sec.}{Secs.}
\Crefname{section}{Section}{Sections}
\Crefname{table}{Table}{Tables}
\crefname{table}{Tab.}{Tabs.}

\def\cvprPaperID{7834} 
\def\confName{CVPR}
\def\confYear{2023}

\begin{document}

\title{Towards Robust Video Instance Segmentation with Temporal-Aware Transformer}

\author{Zhenghao Zhang
,
Fangtao Shao
,
Zuozhuo Dai 
,
Siyu Zhu\\
Alibaba Group\\
{\tt\small \{zhangzhenghao.zzh,shaofangtao.sft,zuozhuo.dzz,siting.zsy\}@alibaba-inc.com}
}
\maketitle
\begin{abstract}
Most existing transformer based video instance segmentation methods extract per frame features independently, hence it is challenging to solve the appearance deformation problem. In this paper, we observe the temporal information is important as well and we propose TAFormer to aggregate spatio-temporal features both in transformer encoder and decoder. Specifically, in transformer encoder, we propose a novel spatio-temporal joint multi-scale deformable attention module which dynamically integrates the spatial and temporal information to obtain enriched spatio-temporal features. In transformer decoder, we introduce a temporal self-attention module to enhance the frame level box queries with the temporal relation. Moreover, TAFormer adopts an instance level contrastive loss to increase the discriminability of instance query embeddings. Therefore the tracking error caused by visually similar instances can be decreased. Experimental results show that TAFormer effectively leverages the spatial and temporal information to obtain context-aware feature representation and outperforms state-of-the-art methods.
\end{abstract}
\section{Introduction}
\label{sec:intro}

\begin{figure}[!t]
\centering
\includegraphics[width=3.2in]{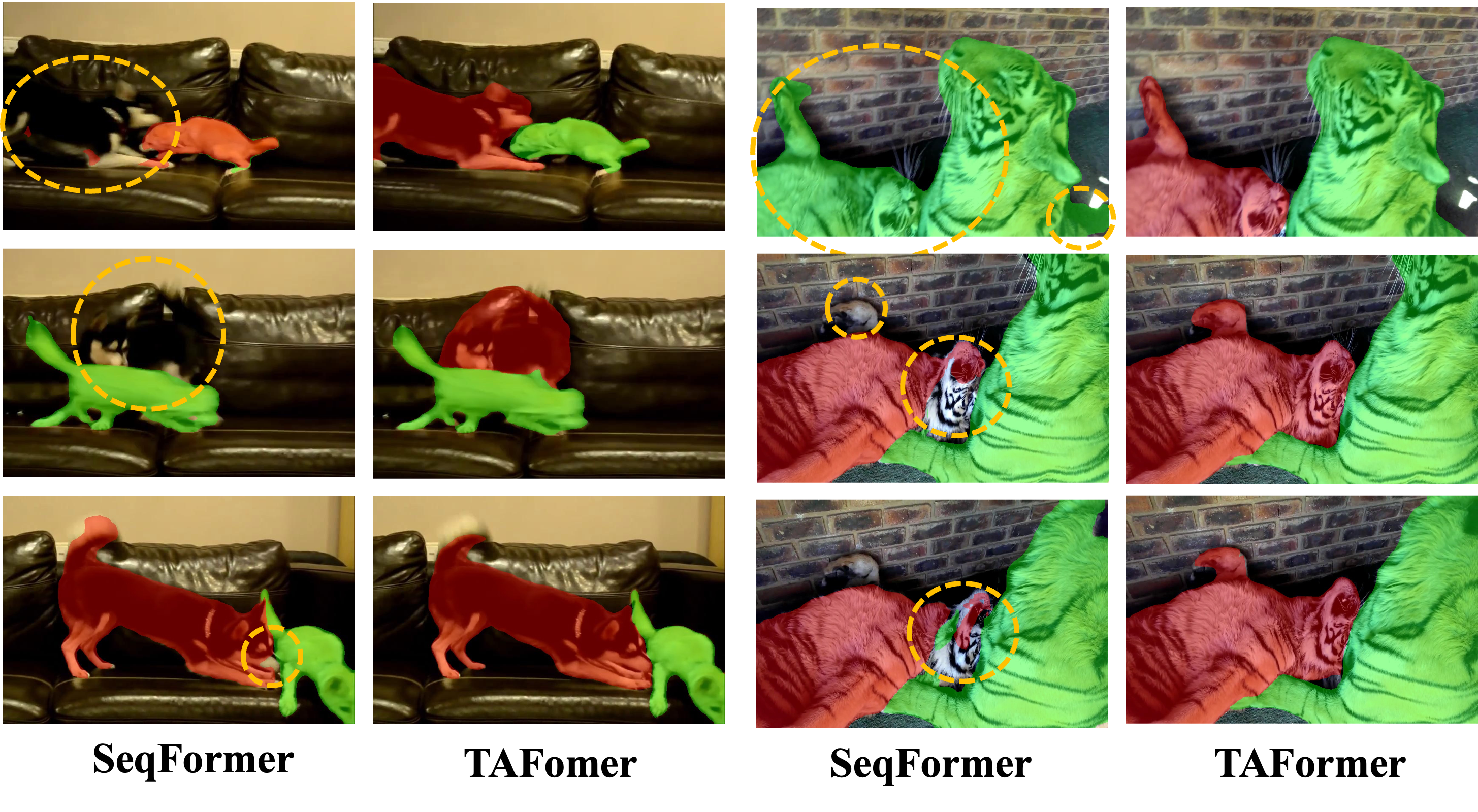}
\caption{Qualitative comparison of the proposed TAFormer with SeqFormer in the cases of appearance deformations and similar object interference. For the first two columns, SeqFormer struggles to accurately segment the instance of the black dog (marked by yellow circles) because of fast motion and scale variation. For the last two columns, it is difficult for SeqFormer to maintain the consistency of the tiger class due to the visual similarity and occlusions. Our TAFormer alleviates these two issues and remarkably improves the mask quality. Best viewed zoomed in.
}
\label{Fig. 1}
\vspace{-3mm}
\end{figure}

Video instance segmentation (VIS) is firstly proposed in the work~\cite{yang2019video} and gradually becomes an emerging task in computer vision.
The goal of VIS is to simultaneously perform detection, segmentation and tracking of the instances in the given videos.
Generally speaking, there are mainly two kinds of approaches.
First, the classic methods known as tracking-by-detection~\cite{bertasius2020classifying,yang2019video,cao2020sipmask,fu2021compfeat,liu2021sg,li2021spatial,yang2021crossover,xu2021segment} apply the standard image instance segmentation approaches~\cite{he2017mask, wang2020solo, wang2020solov2} to the video frame by frame.
A post-process step is applied to
track instances across frames by motion prediction~\cite{bertasius2020classifying}, classification~\cite{qi2021occluded}, and re-identification~\cite{cao2020sipmask, li2021spatial}.
However, such a tracking-by-detection paradigm is sensitive to appearance deformation such as occlusions, motion blur, and large variance in object scales.
The second branch of approaches~\cite{wang2021end,hwang2021video, wu2021seqformer,cheng2021mask2former,yang2022temporally} divide the whole video into multiple clips with overlapping and process the video clip by clip.

Thanks to the emerging advances in vision transformer architecture~\cite{carion2020end,zhu2020deformable,sun2021rethinking,liu2021swin}, recent transformer based VIS works~\cite{wang2021end,hwang2021video, wu2021seqformer,cheng2021mask2former,yang2022temporally} follow the second clip-level paradigm and represent each instance as a learned query embedding.
Specifically, VisTR~\cite{wang2021end} is the first approach that applies transformer to VIS.
It processes the video clip and directly regards the task of VIS as an end-to-end parallel sequence prediction problem~\cite{carion2020end}.
Subsequently, IFC~\cite{hwang2021video} proposes an inter-frame communication transformer and remarkably improves the computation and memory efficiency of frame-to-frame communications.
Moreover, TeViT~\cite{yang2022temporally} proposes a nearly convolution-free and temporally efficient vision transformer.
In SeqFormer~\cite{wu2021seqformer}, the instance of each frame is located and the corresponding salient representation of video-level instance is aggregated by temporal information.

Although, the state-of-the-art SeqFormer~\cite{wu2021seqformer} and transformer based methods of the similar paradigm achieve superior performance in VIS, two main challenges (see \cref{Fig. 1}) still exist. 
First, since SeqFormer~\cite{wu2021seqformer} extracts per-frame features independently,
the ignorance of the spatial and temporal information in transformer encoder module probably leads to the problem of appearance deformations due to the fast motion, occlusion, and scale variation. 
Second, the classification loss generally adopted in the image object detection task is not sufficient to discriminate the instances of the same category in the videos especially when the multiple instances are of the same category and visually similar.

In this paper, we propose the Temporal-Aware Transformer (TAFormer) on top of SeqFormer~\cite{wu2021seqformer} and try to alleviate the two problems mentioned above.
To alleviate the appearance deformation issue, in transformer encoder, we extend the deformable attention module~\cite{zhu2020deformable} to the spatio-temporal domain,
and introduce a spatio-temporal joint multi-scale deformable attention (STJ-MSDA) module to aggregate spatio-temporal features.
In the transformer decoder, a standalone instance query is decomposed to frame-level box queries where each of them represents the box location in the corresponding frame.
We observe that the box location in the current frame should be constrained by the corresponding box locations in adjacent frames.
Therefore, we introduce the consistency constraint to enhance the frame level box queries with the temporal relation.
To enhance the discriminability of visually similar instances, we also propose an instance-level contrastive loss on the box queries, where box queries from the same instance are positive samples and the others are negative samples.

We evaluate TAFormer on two benchmark VIS datasets (Youtube-VIS2019~\cite{yang2019video} and Youtube-VIS2021~\cite{yt2021}) with no additional training data. 
With the ResNet50~\cite{he2016deep} backbone, our method achieves 48.1\% $\rm AP$ on Youtube-VIS2019 and 42.1\% $ \rm AP$ on Youtube-VIS2021, surpassing previous state-of-the-art method Mask2Former~\cite{cheng2021mask2former} by up to 2\%. 


\begin{figure*}[!t]
\centering
\includegraphics [width=6.8in]{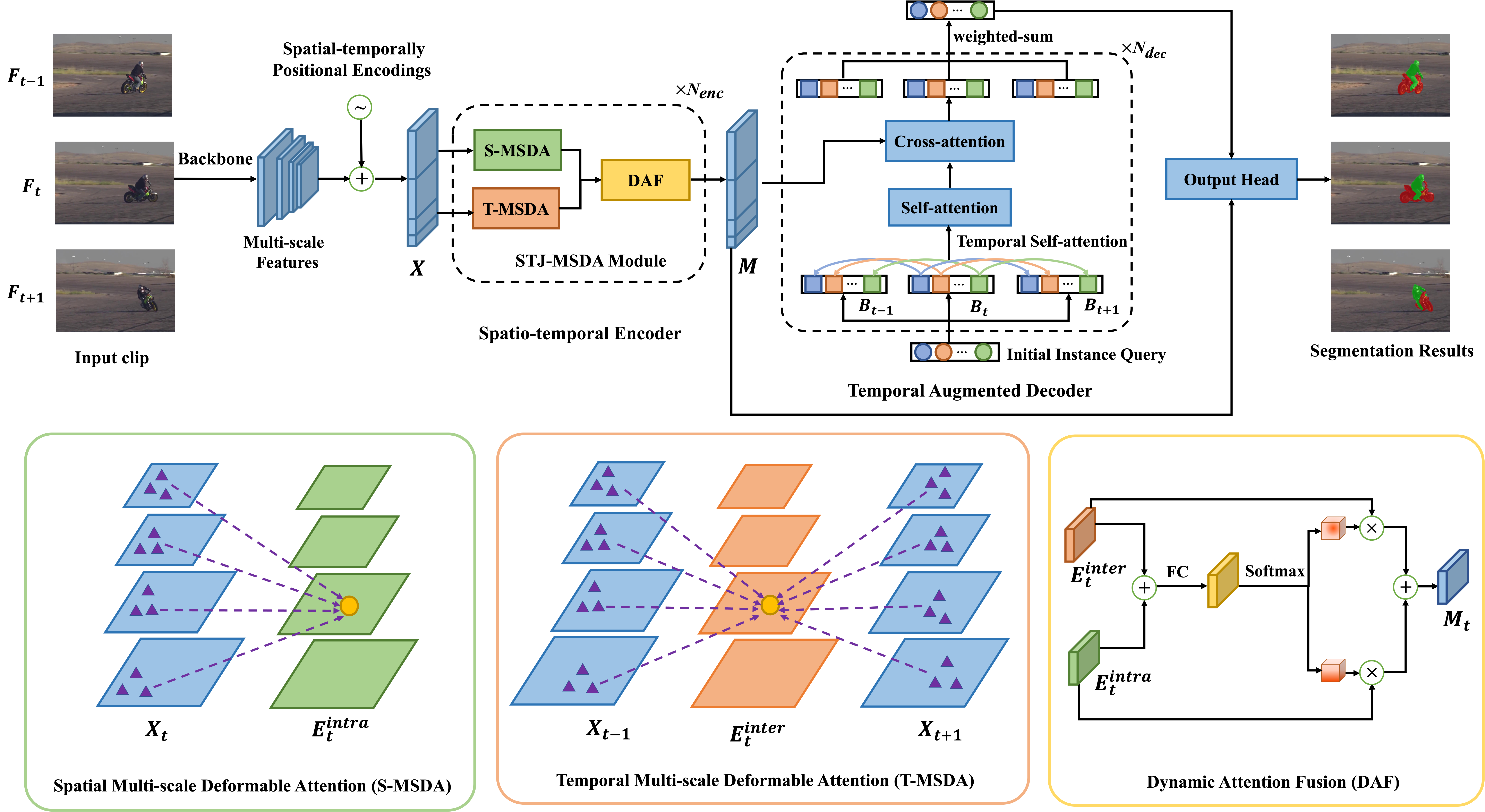}
\caption{The overall illustration of our proposed TAFormer, mainly consisting of the backbone, spatio-temporal encoder, temporal augmented decoder, and output head. More specifically, given the multi-scale feature maps of a video clip, our encoder first extracts spatial and temporal features from the current frame and bidirectional neighboring frames by multi-scale deformable attention. Afterwards, the encoder dynamically integrates the spatial and temporal features to obtain enriched spatio-temporal features. For the temporal augmented decoder, the initial instance query is decomposed into frame-level box queries such that the temporal self-attention is conducted to keep consistency across frames. 
Finally, the output head combines the encoder feature maps with instance queries to generate the sequential  segmentation results. In this figure, the same color represents the box queries belonging to the same video instance.  Best viewed in color.}
\label{Fig. 2}
\vspace{-3mm}
\end{figure*}

\section{Related Works}

\noindent \textbf{Image Instance Segmentation.} Image instance segmentation is a complex computer vision task that combines object detection and semantic segmentation, requiring pixel-level classification and localization of each instance. Image instance segmentation methods can be divided into three groups: two-stage, one-stage, and query-based methods.
Two-stage methods~\cite{he2017mask,cai2018cascade,chen2019hybrid,huang2019mask,kirillov2020pointrend}
first generates regions of interest and then performs regression in each ROI for classification and segmentation. Mask RCNN~\cite{he2017mask} is the representative two-stage detection framework, which is an extension of Fast RCNN~\cite{ren2015faster}. 
One-stage methods~\cite{dai2016instance,li2017fully,chen2019tensormask,chen2020blendmask} get the results directly without proposal generation and the representatives include PolarMask~\cite{xie2020polarmask}, YOLACT~\cite{bolya2019yolact}, SOLO~\cite{wang2020solo} and CondInst~\cite{tian2020conditional}.
Query-based instance segmentation methods such as ISTR~\cite{hu2021istr}, QueryInst~\cite{fang2021instances}, and SOLQ~\cite{dong2021solq} treat the segmentation as a set prediction problem.
These methods use queries to represent the interest objects and perform regression on classification, detection, and mask simultaneously.

\noindent \textbf{Frame Level Video Instance Segmentation.} 
The VIS task aims to classify and segment the instances in each frame and associate the same instance across the whole video. 
VIS methods can be divided into two groups: frame-level methods~\cite{yang2019video,li2021spatial,fu2021compfeat,bertasius2020classifying,cao2020sipmask} and clip-level methods~\cite{athar2020stem,wang2021end,hwang2021video,wu2021seqformer,cheng2021mask2former}. The frame-level methods extend the image instance segmentation model to the VIS task by introducing an additional tracking branch. As a pioneering method, MaskTrack R-CNN~\cite{yang2019video} extends Mask R-CNN~\cite{he2017mask} and introduces an additional tracking head to address the issue of instance association. SipMask~\cite{cao2020sipmask} introduces the one-stage image detector FCOS~\cite{tian2019fcos} and YOLACT~\cite{bolya2020yolact++} to segment instances and associates them in a similar manner. STMask~\cite{li2021spatial} proposes a temporal fusion module to capture temporal correlation, the time-enhanced features are used for segmentation on each frame. CompFeat~\cite{fu2021compfeat} aggregates the temporal and spatial context information to refine the segmentation results in challenging video scenarios. Maskprop\cite{bertasius2020classifying} utilizes the predicted masks generated by the Hybrid Task Cascade Network\cite{chen2019hybrid} and propagates them to other frames for better association. 

\noindent \textbf{Clip Level Video Instance Segmentation.}
The clip-level methods take a video clip as input and generate the sequential segmentation results.
STEm-Seg~\cite{athar2020stem} takes a video clip as a spatio-temporal volume and separates the instances by clustering. 
More recently, there are various clip-level methods built upon the vision transformer~\cite{dosovitskiy2020image,child2019generating,ainslie2020etc,zhu2021long,liu2021swin,zhu2020deformable,zhou2022transvod,wang2022deformable} and making remarkable progress.
As a representative method, 
VisTR~\cite{wang2021end} designs frame-specific instance queries and proposes a new transformer paradigm of joint tracking and segmentation.
However, the dense full attention across the spatio-temporal domain leads to expensive calculations. 
In order to reduce the computation of VisTR, IFC~\cite{hwang2021video} proposes memory tokens in the transformer encoder and adopts video-level object queries for segmentation. 
Later on, Mask2Former~\cite{cheng2021mask2former} and SeqFormer~\cite{wu2021seqformer} improve the performance on the benchmark datasets.
Specifically, Mask2Former~\cite{cheng2021mask2former} introduces a masked-attention within the transformer architecture for instance segmentation.
It trivially generalizes to video segmentation by directly predicting 3D segmentation results.
SeqFormer~\cite{wu2021seqformer} introduces a frame-independent transformer architecture, which decomposes the video-level instance queries to each frame for frame-wise detection and segmentation.
Both of them integrate Deformable DETR~\cite{zhu2020deformable} and introduce multi-scale representations in a low computational cost manner.
But there are only intra-frame multi-scale features utilized in attention computation.
Thereby, they ignore the spatio-temporal correlation across frames, which is crucial to handle challenging scenarios.
Our model is designed to sufficiently exploit the temporal context across frames, which can effectively handle these challenges and leads to more accurate results.

\section{Methods}
\subsection{Overall Architecture}

As illustrated in \cref{Fig. 2}, we propose Temporal-Aware Transformer (TAFormer) for video instance segmentation, which efficiently exploits the temporal correlation within a video clip. Our proposed TAFormer consists of four main modules: a vision backbone, a spatio-temporal transformer encoder, a temporal augmented transformer decoder, and an output head. Herein, let $\left \{  F_i \right \}_{i=t-d}^{t+d}$ be the input clip, which consists of frame $t$ and its bidirectional neighboring frames with a maximum distance of $d$. The backbone network extracts multi-scale features frame by frame. In each stage of the transformer encoder, we utilize a novel spatio-temporal joint deformable attention (STJ-MSDA) as the critical component of our pipeline. In particular, there are three sub-modules in the proposed STJ-MSDA architecture, mainly including spatial multi-scale deformable attention (S-MSDA),  temporal multi-scale deformable attention (T-MSDA) and dynamic attention fusion (DAF). Moreover, we further integrate the temporal self-attention block into the query-decomposed decoder~\cite{wu2021seqformer} to enhance the temporal correlation on frame-level box queries. Finally, we employ a contrastive loss~\cite{Wu2018UnsupervisedFL} over box queries from all frames for robustness during training. 

\subsection{Spatio-temporal Transformer Encoder}
\label{3.2}

Lacking temporal information hinders promising segmentation results in challenging video scenarios. To solve this problem, we propose a Spatio-temporal Transformer Encoder (ST-Enc), which is composed of a stack of STJ-MSDA modules. Different from the original Multi-scale Deformable Attention Module~\cite{zhu2020deformable}, our proposed STJ-MSDA module employs temporal attention from adjacent frames and selectively aggregates it with spatial attention to obtain deeper context-aware representation.
Specifically, given an element query $q$ from frame $t$, STJ-MSDA first attends multi-scale spatial locations in the current frame and neighboring frames, facilitates acquiring highly-independent spatial and temporal deformable attentions.
It then dynamically aggregates information from these two attention via DAF. In this way,  more discriminative spatio-temporal representations can be provided. We elaborate on each component in detail in the following paragraphs.

\noindent \textbf{Spatial Multi-scale Deformable Attention.} Following Deformable DETR~\cite{zhu2020deformable}, 
S-MSDA computes attention on points sampling from the multi-scale feature maps of the current frame. 
We define the $l\text{-}th$ feature map from frame $t$ as  $x^{l}_t \in  \mathbb{R}^{C\times H_{l} \times W_{l}}$ and the normalized coordinates of the reference point for $q$ as $\hat{p} _{q} \in {[0,1]}^2 $. Then given the query feature $z_{q}$, the S-MSDA can be described as follows:
\begin{normalsize}
\begin{equation}
\begin{aligned}
\mathcal{S\text{-}MSDA} & (z _{q}, \hat{p} _{q}, x_{t}^{l}) = \sum_{m=1}^{M}W_{m} \cdot [\sum_{l=1}^{L}\sum_{k=1}^{K_{intra}} A_{mlqk} \\ & \cdot W_{m}^{'} x^{l}_t(\phi_{l}(\hat{p} _{q})+\Delta p_{mlqk} ) ], 
\end{aligned}
\end{equation}
\end{normalsize}
where $m$ denotes the index of multiple attention heads , $K_{intra}$ is the total sampled number from each feature level. $\Delta p_{mlqk}$ and $A_{mlqk}$ represent the pixel offset and attention weight of the $k\text{-}th$ sampling point in the $l\text{-}th$ feature level. Both of them are obtained via linear projection over $z_q$. The attention weight is normalized over all sampled points by $\sum_{l=1}^{L}\sum_{k=1}^{K_{intra}}A_{mlqk}=1$. Function $\phi_{l}$ re-scales the normalized coordinates $\hat{p} _{q}$ to the input feature map of the $l\text{-}th$ level.

\noindent \textbf{Temporal Multi-scale Deformable Attention.} We extend S-MSDA to the temporal domain and propose T-MSDA. The core idea of the T-MSDA is that, for each query in the current frame, it attends to a small set of key sampled points from multi-scale feature maps of neighboring frames. As such, our method can aggregate enriched temporal representation and maintain temporal consistency of the foreground instances. It can be formulated as follows:
\begin{normalsize}
\begin{equation}
\begin{aligned}
\mathcal{T\text{-}MSDA} & (z _{q}, \hat{p} _{q}, x_{t'}^{l}) = \sum_{m=1}^{M}W_{m} \cdot [\sum_{t'=t-d \atop t'!=t}^{t+d}\sum_{l=1}^{L}\sum_{k=1}^{K_{inter}} \\ & A_{mltqk}  \cdot W_{m}^{'} x_{t'}^{l} (\phi_{l}(\hat{p} _{q})+\Delta p_{mtlqk} ) ],
\end{aligned}
\end{equation}
\end{normalsize}
where $t'$ denotes the neighboring frame with frame $t$. $K_{inter}$ represents the number of sampled points in each feature level across neighboring frames. The scaled attention weight is normalized  by $\sum_{l=1}^{L}\sum_{t'=t-d}^{t+d}\sum_{k=1}^{K_{inter}}A_{mltqk}=1$. $p_{mtlqk}$ and $A_{mltqk}$ are obtained by two different linear projection over $z_q$, respectively.


\noindent \textbf{Dynamic Attention Fusion.} Let $E_{t}^{intra}$ and $E_{t}^{inter}$ represent the attention maps for frame $t$ obtained by $\mathcal{S\text{-}MSDA}$ and $\mathcal{T\text{-}MSDA}$, respectively. To our knowledge, $E_{t}^{intra}$ contains more salient instance information in the current frame, but struggles to handle the appearance deformations across frames. In contrast, $E_{t}^{inter}$ contains deeper contexts in order to adapt to the appearance changes. Therefore, the spatial and temporal attention play different roles in various video scenes. Based on the analysis above, we propose the Dynamic Attention Fusion module to emphasize the semantic information with channel-wise attention. We first aggregate them by an element-wise addition as:
\begin{normalsize}
\begin{equation}
\begin{aligned}
E'=E_{t}^{intra} + E_{t}^{inter}.
\end{aligned}
\end{equation}
\end{normalsize} 
Then, we apply a fully connected (FC) layer to achieve the adaptive selection from $E'$. After that, we employ another two FC layers to generate the gate vectors $\left \{  g_1,g_2\right \} $, such that the information flows from these two attention maps $\left \{  E_{t}^{intra}, E_{t}^{inter}\right \} $ can be controlled. We further utilize softmax function along the channel dimension to generate adaptive weights $\left \{  w_1,w_2\right \}$ as follows:
\begin{normalsize}
 \begin{equation}
\begin{aligned}
w_1=\frac{e^{g_{1}}}{e^{g_{1}} + e^{g_{2}}} \quad  w_2=\frac{e^{g_{2}}}{e^{g_{1}} + e^{g_{2}}}.
\end{aligned}
\end{equation}
\end{normalsize}
Finally, the enriched spatio-temporal features of frame $t$ can be obtained by weighted addition:
\begin{normalsize}
\begin{equation}
\begin{aligned}
M_{t} = E_{t}^{intra} \odot w_1 + E_{t}^{inter} \odot w_2.
\end{aligned}
\end{equation}
\end{normalsize}

\subsection{Spatio-temporal Positional Encoding}
Different from SeqFormer that only employs spatial positional encoding, we further extend the positional encoding to the spatio-temporal domain, which can be described as follows:
\begin{normalsize}
\begin{equation}
\begin{aligned}
e_{pos} = e_{pos}^{t} +  e_{pos}^{s},
\end{aligned}
\end{equation}
\end{normalsize} where $e_{pos}^{s} \in  \mathbb{R}^{C\times 1\times H_{l} \times  W_{l}}$ denotes the original spatial encoding in SeqFormer and $e_{pos}^{t} \in  \mathbb{R}^{C\times T\times 1 \times 1}$ represents the corresponding temporal positional encoding. Both $e_{pos}^{s}$ and $e_{pos}^{t}$ are obtained by sine and cosine functions of different frequencies. With spatio-temporal positional encoding, we can deal with arbitrary length video clip.

\subsection{Temporal Augmented Transformer Decoder}
\label{3.4}

Following the SeqFormer architecture that decomposes the instance query into frame-level box queries and aggregates the box queries to obtain a temporal-aware instance query, we further employ a Temporal Self-Attention (TSA) block to enhance the temporal relationship. More specifically, the Temporal Self-Attention computes attention on box queries belonging to the same instance but different frames to obtain temporal-aware box queries.
These box queries are then passed through the self-attention module and cross-attention module to query the enriched features from ST-Enc, thus accurately exploring location information.


\subsection{Contrastive Loss on Box Queries}
\label{3.5}
Our encoder aims to produce enriched spatio-temporal features, whilst the decoder focuses on improving the temporal consistency for predictions. As such, the problem of appearance deformation in video instance segmentation can be efficiently solved. However, considering the challenging scenarios with multiple visual similar instances, it is easy to accumulate association errors over time due to the non-discriminative embeddings among queries. Motivated by the success of contrastive learning~\cite{Wu2018UnsupervisedFL,he2020momentum}, we propose a temporal contrastive loss to map different instances box queries to distinct representations. For each video clip, this loss regards the box queries belonging to the same instance as positive samples and all other box queries as negative samples. 

\begin{figure}[!t]
\includegraphics [width=3in]{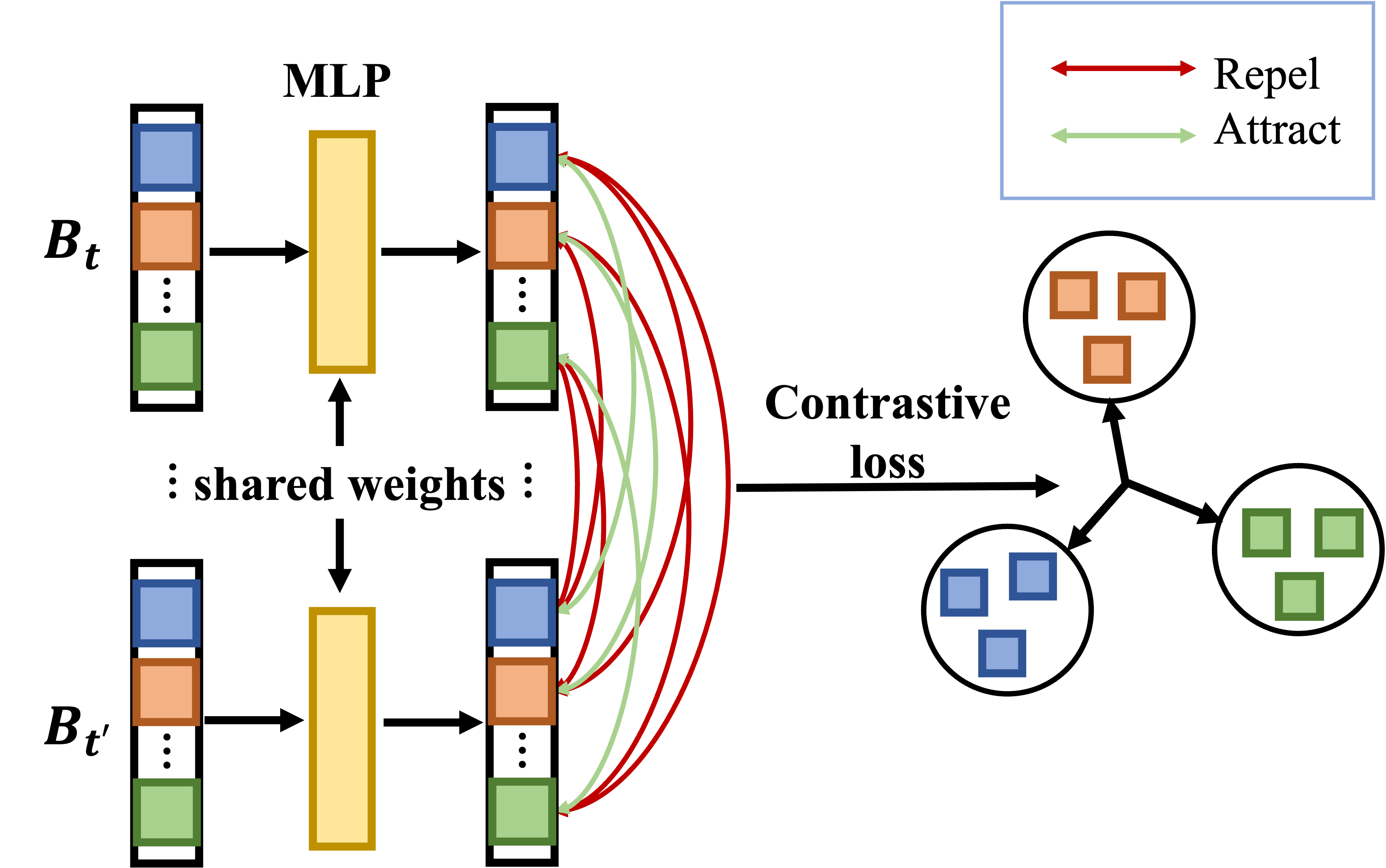}
\caption{Contrastive learning on frame-level box queries.}
\vspace{-3mm}
\label{Fig. 3}
\end{figure}

As illustrated in \cref{Fig. 3}, we define $B_t={\left \{  b_t^q \right \} }_{q=1}^{Q}$ be the box queries of frame $t$, where $b_t^q\in R^C$ and $Q$ is the total number of queries. The core idea of contrastive learning is to maximize agreement between differently augmented views of the same example in the latent space. Automatically, ${\left \{  b_{t}^{i}, b_{t'}^{i} \right \} }$ from frame $t$ and its neighboring frame $t'$ can be treated as the different views of the $i\text{-}th$ instance in terms of temporal domain. We first adopt two FC layers to generate the latent representations of box queries. Then, InfoNCE~\cite{oord2018representation}~ is employed between two frames. It focuses on maximizing the representations between the pairs of box queries with the same order and minimizing the similarity with other pairs, which can be described as follows:
\begin{normalsize}
\begin{equation}
\begin{aligned}
\mathcal{L}_{N}(B_{t}, B_{t'})=-\frac{1}{Q}\sum_{i}^{Q}log\frac{exp(s(b_{t}^{i}, b_{t'}^{i})/\tau )}{\sum_{j=1}^{Q}exp(s(b_{t}^{i}, b_{t'}^{j})/\tau)},
\end{aligned}
\end{equation}
\end{normalsize} where $\tau$ is temperature hyper-parameter and the similarity function $s(b_{i}, b_{j})$ is defined as the cosine similarity. We further compute $\mathcal{L}_{N}$ between all pairs cross frames to obtain the contrastive loss, which can be formulated by:

\begin{normalsize}
\begin{equation}
\begin{aligned}
\mathcal{L}_{cl}=\sum_{t}^{T}\sum_{t'\ne t}^{T}\mathcal{L}(B_{t},B_{t'}).
\end{aligned}
\end{equation}
\end{normalsize}
The total losses for TAFormer can be defined as follows:
\begin{normalsize}
\begin{equation}
\begin{aligned}
\mathcal{L}_{total} = \lambda_{cls}\cdot\mathcal{L}_{cls} + \mathcal{L}_{box} + \mathcal{L}_{mask} + \mathcal{L}_{cl},
\end{aligned}
\end{equation}
\end{normalsize} where $\mathcal{L}_{cls}$, $\mathcal{L}_{box}$ and $\mathcal{L}_{mask}$ represent classification loss, bounding box loss, and segmentation mask loss. Herein, we use focal loss~\cite{lin2017focal} as the classification loss. The box loss consists of L1 loss and GIou loss~\cite{rezatofighi2019generalized}: $\mathcal{L}_{box} = \lambda_{L1}\cdot\mathcal{L}_{L1} + \lambda_{giou}\cdot\mathcal{L}_{giou} $. The mask loss is formulated as the combination of dice loss~\cite{MilletariNA16} and binary mask focal loss:~ $\mathcal{L}_{mask} = \lambda_{dice}\cdot\mathcal{L}_{dice} + \lambda_{focal}\cdot\mathcal{L}_{focal} $. Following~\cite{zhu2020deformable, wang2021end, hwang2021video, wu2021seqformer, cheng2021mask2former}, we apply auxiliary losses~\cite{al2019character} on box queries after each decoding layer.

\begin{table*}[h]\small
\centering
\renewcommand{\arraystretch}{1.1}
\begin{tabular}{lcccccccc}
\hline
\multicolumn{1}{c}{backbone}                     & Methods          & Volume    & Params & $\rm AP$   & $\rm AP_{50}$ & $\rm AP_{75}$ & $\rm AR_{1}$  & $\rm AR_{10}$ \\ \hline
\multicolumn{1}{l|}{} & MaskTrack R-CNN~\cite{yang2019video}  & ICCV19    & 58.1M  & 30.3 & 51.1 & 32.6 & 31.0 & 35.5 \\
\multicolumn{1}{l|}{}                            & STEm-Seg~\cite{athar2020stem}         & ECCV20    & 50.5M  & 30.6 & 50.7 & 33.5 & 37.6 & 37.1 \\
\multicolumn{1}{l|}{}                            & SipMask~\cite{cao2020sipmask}          & ECCV20    & \textbf{33.2}M  & 33.7 & 54.1 & 35.8 & 35.4 & 40.1 \\
\multicolumn{1}{l|}{}                            & CompFeat~\cite{fu2021compfeat}         & AAAI21    & -      & 35.3 & 56.0 & 38.6 & 33.1 & 40.3 \\
\multicolumn{1}{l|}{}                            & VisTR~\cite{wang2021end}            & CVPR21    & 57.2M  & 36.2 & 59.8 & 36.9 & 37.2 & 42.4 \\
\multicolumn{1}{l|}{}                            & MaskProp~\cite{bertasius2020classifying}         & CVPR20    & -      & 40.0 & -    & 42.9 & -    & -    \\
\multicolumn{1}{l|}{{ResNet-50}}                & CrossVIS~\cite{yang2021crossover}         & ICCV21    & 37.5M  & 36.3 & 56.8 & 38.9 & 35.6 & 40.7 \\
\multicolumn{1}{l|}{}                            & Propose-Reduce~\cite{lin2021video}   & ICCV21    & 69.0M  & 40.4 & 63.0 & 43.8 & 41.1 & 49.7 \\
\multicolumn{1}{l|}{}                            & PCAN~\cite{KeLDTTY21}             & NeurIPS21 &  -     & 36.1 &54.9& 39.4 & 36.3 & 41.6 \\
\multicolumn{1}{l|}{}                            & IFC~\cite{hwang2021video}              & NeurIPS21 & 39.3M  & 42.8 & 65.8 & 46.8 & 43.8 & 51.2 \\
\multicolumn{1}{l|}{}                            & Mask2Former~\cite{cheng2021mask2former}      & CVPR22    & -      & 46.4 & 68.0 & 50.0 & -    & -    \\
\multicolumn{1}{l|}{}                            & SeqFormer~\cite{wu2021seqformer}        & ECCV22    & 49.3M  & 45.1 & 66.9 & 50.5 & 45.6 & 54.6 \\
\multicolumn{1}{l|}{}                            & \textbf{TAFormer} & -         & 52.2M  & \textbf{48.1} & \textbf{71.6} & \textbf{52.5} & \textbf{45.9} & \textbf{56.6} \\ \hline
\multicolumn{1}{l|}{}            & CrossVIS~\cite{yang2021crossover}        & ICCV21    & 56.6M  & 36.6 & 57.3 & 39.7 & 34.4 & 41.6 \\
\multicolumn{1}{l|}{}                             & PCAN~\cite{KeLDTTY21}              & NeurIPS21 & -      & 37.6 & 57.2 & 41.3 & 37.2 & 43.9 \\
\multicolumn{1}{l|}{}                             & VisTR~\cite{wang2021end}            & CVPR21    & 76.3M  & 38.6 & 61.3 & 42.3 & 37.6 & 44.2 \\
\multicolumn{1}{l|}{ResNet-101}                             & SeqFormer$\ddagger$~\cite{wu2021seqformer}        & ECCV22    & 68.4M  & 49.0 & 71.1 & 55.7 & \textbf{46.8} & 56.9 \\
\multicolumn{1}{l|}{}                             & Mask2Former~\cite{cheng2021mask2former}      & CVPR22    & -      & 49.2 & 72.8 & 54.2 & -    & -    \\
\multicolumn{1}{l|}{}                             & \textbf{TAFormer}           & -         & 71.3M  & \textbf{50.6} & \textbf{74.2} & \textbf{57.4} & 46.0 & \textbf{58.1} \\ \hline
\multicolumn{1}{l|}{}            & SeqFormer$\ddagger$~\cite{wu2021seqformer}        & ECCV22    & 220.0M  & 59.3 & 82.1 & 66.4 & 51.7 & 64.4 \\
\multicolumn{1}{l|}{Swin-L}                               & Mask2Former~\cite{cheng2021mask2former}              & CVPR22 & -      & 60.4 & 84.4 & 67.0 & - & - \\
\multicolumn{1}{l|}{}                             & \textbf{TAFormer}            & -    & 222.9M  & \textbf{61.2} & \textbf{85.0} & \textbf{67.4} & \textbf{54.6} & \textbf{66.7} \\ \hline
\end{tabular}
\caption{Performance of our method compared to state of the art on YTVIS-19 validation set. $\ddagger$ denotes adopting 80K train2017 images containing YTVIS-19 categories for joint training.}
\label{Tab1}
\end{table*}

\begin{table*}[h]\small
\centering
\renewcommand{\arraystretch}{1.1}
\begin{tabular}{ccccccccc}
\hline
Methods           & Backbone & Volume    & Params & $\rm AP$   & $\rm AP_{50}$          & $\rm AP_{75}$          & $\rm AR_{1}$  & $\rm AR_{10}$ \\ \hline
VisTR~\cite{wang2021end}          & ResNet50 & CVPR21    & 57.2M  & 31.8 & 51.7          & 34.5          & 29.7 & 36.9 \\
IFC~\cite{hwang2021video}               & ResNet50 & NeurIPS21 & \textbf{39.3M}  & 36.6 & 57.9          & 39.3          & -    & -    \\
Mask2Former~\cite{cheng2021mask2former}        & ResNet50 & CVPR22    & -      & 40.6 & 60.9          & 41.8          & -    & -    \\
SeqFormer$\ddagger$~\cite{wu2021seqformer}          & ResNet50 & ECCV22    & 49.3M  & 40.5 & 62.5          & 43.6          & 36.2 & 48.0 \\
TeViT~\cite{yang2022temporally}             & MsgShifT & CVPR22    & 172.3M & 37.9 & 61.2          & 42.1          & 35.1 & 44.6 \\
\textbf{TAFormer} & ResNet50 & -         & 52.2M  & \textbf{42.1} & \textbf{63.8} & \textbf{44.8} & \textbf{38.6} & \textbf{49.2} \\ \hline
\end{tabular}
\caption{Quantitative results of VIS methods on the YTVIS-21 validation set. $\ddagger$ denotes adopting 80K train2017 images containing YTVIS-21 categories for joint training.}
\label{Tab2}
\vspace{-3mm}
\end{table*}

\section{Experiments}
\subsection{Experimental Setup}
\noindent \textbf{Datasets.} We evaluate our TAFormer on both Youtube-VIS2019 (YTVIS-19) and Youtube-VIS2021 (YTVIS-21) datasets. The YTVIS-19 dataset consists of 2238 training, 302 validation, and 343 test videos. Each video has been manually annotated in pixel-level and the semantic category number is 40. The YTVIS-21 dataset adds more samples on the basis of YTVIS-19 with 3859 high-quality videos, whilst the number of annotations is also doubled so that it is more difficult in multi-instance scenarios.

\noindent \textbf{Evaluation Metrics.} We adopt Average Precision (AP) and Average Recall (AR) to evaluate the performance of our model. In addition, the Intersection over Union (IoU) computation is also performed in the whole video.

\noindent \textbf{Implements Details.}  We choose ResNet-50~\cite{he2016deep} as our default backbone network unless otherwise specified. Similar to~\cite{zhu2020deformable}, the feature maps $\left \{ C_3, C_4, C_5 \right \} $ with resolution $1/8$, $1/16$ and $1/32$ from the backbone are utilized to obtain multi-scale information. We also obtain the feature $C_6$ with resolution $1/64$ via a convolution on $C_5$. These four features are concatenated along channel dimensions and sent to our transformer. We set encoder layers $N_{enc} = 6$, decoder layers $N_{dec} = 6$ and hidden dimension $C=256$. Both sampling key numbers $K_{intra}$ and $K_{inter}$ are set to 4 for our STJ-MSDA module. The number of instance queries is set to 300. The coefficients for losses are set as $\mathcal{\lambda}_{cls} = 2$, $\mathcal{\lambda}_{L1} = 5$, $\mathcal{\lambda}_{giou} = 2$, $\mathcal{\lambda}_{dice} = 5$, $\mathcal{\lambda}_{focal} = 2$.

Following ~\cite{wu2021seqformer}, we use the AdamW~\cite{DBLP:conf/iclr/LoshchilovH19} optimizer with a base learning rate of $1 \times 10^{-4}$, $\beta_{1}=0.9$, $\beta_{2}=0.999$, and weight decay $=10^{-4}$. We set the length of video clip $T = 5$ and the temporal sampling windows $d$ to 2 during training phase. Similar to~\cite{athar2020stem , lin2021video, wu2021seqformer}, we first pretrain our model on COCO for 24 epochs where each image is randomly rotated five times with $\pm 20$ degree as a pseudo video clip. Next, we finetune it on YTVIS dataset for 6 epochs. 
The learning rate is reduced by a factor of 0.1 at 3th and 5th epochs during finetune training. Our model is trained on 4 NVIDIA A100 GPUs with 4 video clips per GPU. During reference, we input the down-scaled 360p video as a whole to the model. Therefore, no additional data association is required.

\begin{figure*}[!t]
\centering
\includegraphics [width=6in]{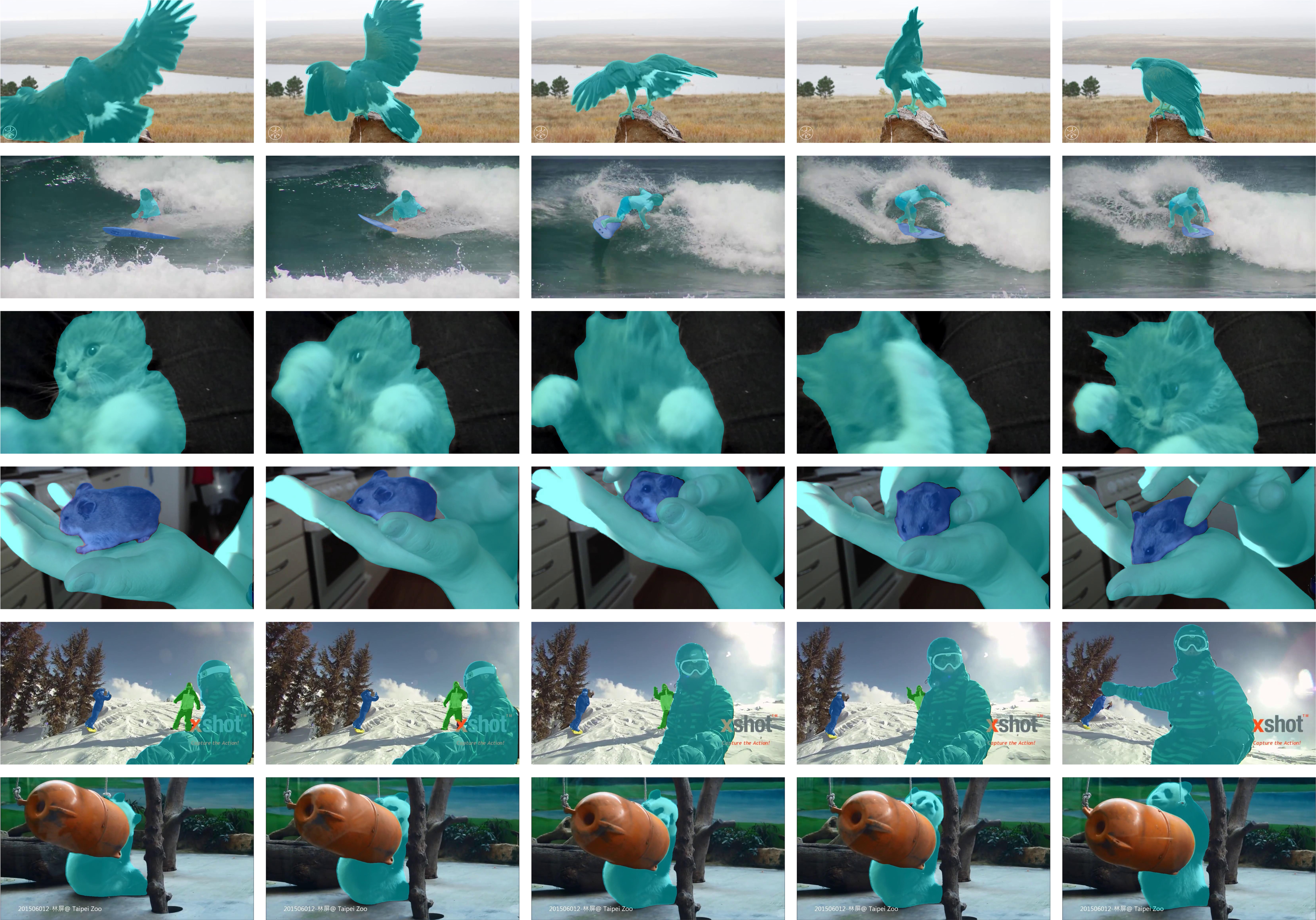}
\caption{Qualitative results of our TAFormer on the YTVIS-19 validation dataset.}
\label{Fig. 4}
\end{figure*}

\subsection{Main Results}

We compare TAFormer with state-of-the-art VIS methods on YTVIS-19 in \cref{Tab1}. SeqFormer and Mask2Former take a strong transformer architecture and achieve promising results. However, both of them only extract multi-scale spatial features and ignore the temporal consistency across frames. Among these competitive methods, our TAFormer achieves a new state-of-art of the 48.1$\% \rm AP$, using the same backbone and training data. Specifically, it provides an absolute gain of 4.7$\%$ at the threshold of $\rm AP_{50}$ over SeqFormer. The higher $ \rm AP $ and $ \rm AR$ validate that our method is able to utilize enrich spatio-temporal information to handle complex scenes and optimize the segmentation results. With the backbone of ResNet-101, our method achieves the overall $\rm AP$ of 50.6$\%$.

\cref{Tab2} illustrates the results of all compared methods on YTVIS-21. Our approach brings an improvement of 1.5$\%$ on $\rm AP$ over the second best method Mask2Former. It is also worth noting that TAFormer outperforms Mask2Former by 3$\%$ at the overlap threshold of $\rm AP_{75}$. It indicates that TAFormer can generate stable trajectories in longer video sequences.

\cref{Fig. 4} visualizes the results of TAFormer in six challenging scenarios. It shows TAFormer can tackle these scenarios well and maintain stable tracking trajectories.


\subsection{Ablation Study}

In this section, we use SeqFormer as our baseline and conduct several ablation studies on it to demonstrate the effectiveness of our proposed method.  All models are evaluated on the YTVIS-19 validation dataset. 
\begin{table}[!t]\small
\centering
\renewcommand{\arraystretch}{1.1}
\begin{tabular}{l|ccc}
\hline
Model                           & $\rm AP$   & $\rm AP_{50}$  & $ \rm AP_{75}$ \\  \hline
Baseline                        & 44.9 & 66.8 & 50.3\\
Baseline + ST-Enc            &  46.4  &68.7 & 51.5\\
Baseline + ST-Enc + TA-Dec & 47.0   & 69.5  & \textbf{52.7}\\ 
Baseline + ST-Enc + TA-Dec + CL & \textbf{48.1} &\textbf{71.6} & 52.5\\ \hline
\end{tabular}
\caption{Ablation Study of our contributions on YTVIS-19 validation dataset. ST-Enc, TA-Dec, and CL denote the
proposed spatio-temporal encoder, temporal augmented decoder, and contrastive loss on box queries, respectively.}
\vspace{-3mm}
\label{Tab3}
\end{table}

\begin{figure}[!t]
\centering
\includegraphics[width=3.2in]{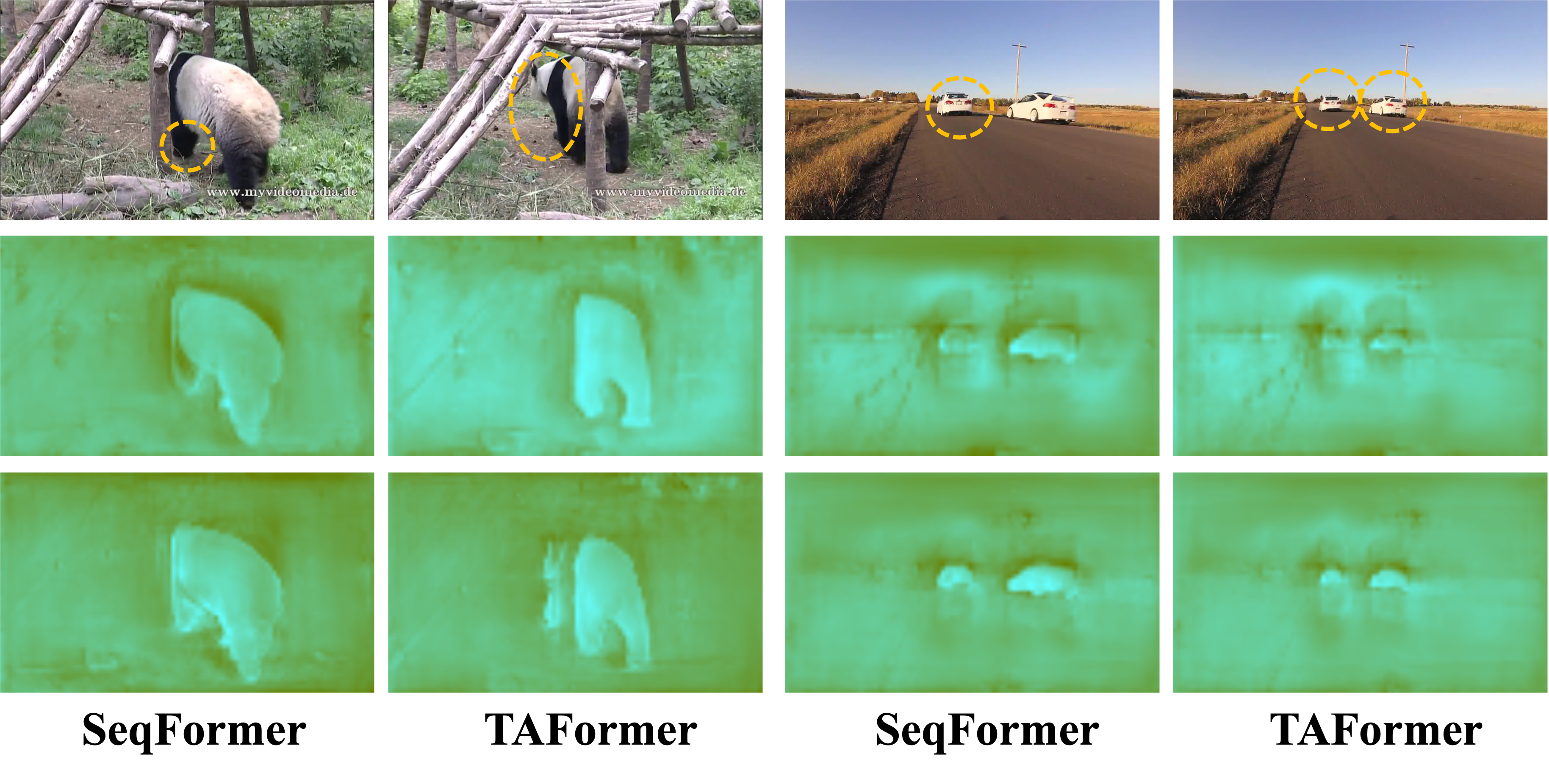}
\caption{Visualization of the feature maps obtained by the encoder. We choose two difficult scenarios including occlusion and scale change. It can be clearly seen that our encoder enhances the regions of foregrounds and suppress the noise of backgrounds. The feature map is generated by average pooling along the channel dimension at the first-scale feature with the size of $H/4$ and $W/4$.}
\label{Fig. 5}
\end{figure}

\noindent \textbf{Spatio-temporal encoder.} We first verify the effectiveness of the spatio-temporal encoder. 
The results are listed in \cref{Tab3}. 
Compared to the baseline model, ST-Enc provides an absolute gain of 1.5 $\%$ $\rm AP$. 
It proves that our encoder has the ability to aggregate meaningful and scalable context features from both intra and inter frames. 
We further visualize the feature maps obtained by the baseline encoder and our encoder in \cref{Fig. 5}. It shows our encoder extracts better foreground regions.

Besides, we evaluate different schemes to aggregate the multi-scale deformable attention from intra-frame and inter-frame. As shown in \cref{Tab4}, the first variant simply combines $E_{t}^{intra}$ with $E_{t}^{inter}$ in an element-wise addition manner. The second variant concatenating $E_{t}^{intra}$ and $E_{t}^{inter}$ brings the $0.3\%$ $\rm AP$ improvement. Our dynamic attention fusion module further boosts the first variant by $0.5\%$ $\rm AP$. These experimental results indicate that our method can provide more discriminative embeddings from different frames and scales for the final segmentation. Thus, we adopt this strategy as the default setting for all experiments.

\begin{table}[!t]\small
\centering
\renewcommand{\arraystretch}{1.1}
\begin{tabular}{l|ccc}
\hline
Fusion strategies  & $\rm AP$   & $\rm AP_{50}$  & $ \rm AP_{75}$ \\  \hline
Element-wise addition                & 45.9 & 67.1 & 51.1\\
Concatenation               &           46.2 & 68.2 & 50.9\\
Dynamic attention fusion               &  \textbf{46.4}  &\textbf{68.7} & \textbf{51.5}\\ \hline
\end{tabular}
\caption{Variants of spatio-temporal attention aggregation. Our dynamic attention fusion achieves a relatively stable performance among all approaches.}
\vspace{-3mm}
\label{Tab4}
\end{table}

In \cref{Tab5}, we study the impact of feature levels on overall performance. We first use a single-scale feature map for deformable attention on the baseline, which is the same as previous transformer-based methods~\cite{wang2021end}. We convert the pretrained weight provided in~\cite{wu2021seqformer} adaptively and get $41.7\%$ AP. The use of temporal sampling on single-scale features can improve $\rm AP$ to $42.6\%$. The multi-scale deformable attention within intra-frame can bring an additional gain of $3.2\%$ on the baseline, which is adopted in MaskFormer and SeqFormer. When integrating temporal multi-scale sampling and dynamic attention module into it, there is a more significant gain ($1.5\%$) on overall $\rm AP$. It proves that the attention module proposed in our encoder leads to more accurate segmentation results since more context information can be utilized.

\begin{table}[tb]\small
\centering
\renewcommand{\arraystretch}{1.1}
\begin{tabular}{c|ccc}
\hline
Model                                          & SFL & TFL & $\rm AP$   \\ \hline
\multicolumn{1}{c|}{\multirow{2}{*}{Baseline}} & 1   & 0   & 41.7 \\
\multicolumn{1}{c|}{}                          & 4   & 0   & 44.9 \\ \hline
\multicolumn{1}{c|}{\multirow{2}{*}{TAFormer}} & 1   & 1   & 42.6 \\
\multicolumn{1}{c|}{}                          & 4   & 4   & \textbf{46.4} \\ \hline
\end{tabular}
\caption{Impact of the feature levels used for sampling points prediction. SFL and TFL denote the spatial feature levels and temporal feature levels, respectively. }
\vspace{-3mm}
\label{Tab5}
\end{table}


\noindent \textbf{Temporal augmented decoder.} As shown in \cref{Tab3}, by performing temporal self-attention before spatial self-attention and deformable cross attention for box queries, it further brings the $ \rm AP$ to $47.0\%$. It is noticed that the performance gain becomes $1.2\%$ in higher overlap $\rm{AP}_{75}$. The reason is that temporal-enhanced box queries can attend more accurate locations with the movement of instances. 

\begin{table}[h]\small
\centering
\renewcommand{\arraystretch}{1.1}
\begin{tabular}{ccccc}
\hline
CL       & Aux     & $\rm AP$ & $\rm AP_{50}$ & $\rm AP_{75}$ \\ \hline
             &              & 47.0 & 69.5         & \textbf{52.7}      \\
$\checkmark$ &              & 47.7 & 71.0         & 51.6      \\
$\checkmark$ & $\checkmark$ & \textbf{48.1} & \textbf{71.6}         & 52.5      \\ \hline
\end{tabular}
\caption{Impact of adding the contrastive loss. Aux denotes adding the auxiliary losses for contrastive learning.}
\vspace{-3mm}
\label{Tab6}
\end{table}

\begin{figure}[!t]
    \begin{minipage}[t]{0.5\linewidth}
        \centering
        \includegraphics[width=\textwidth]{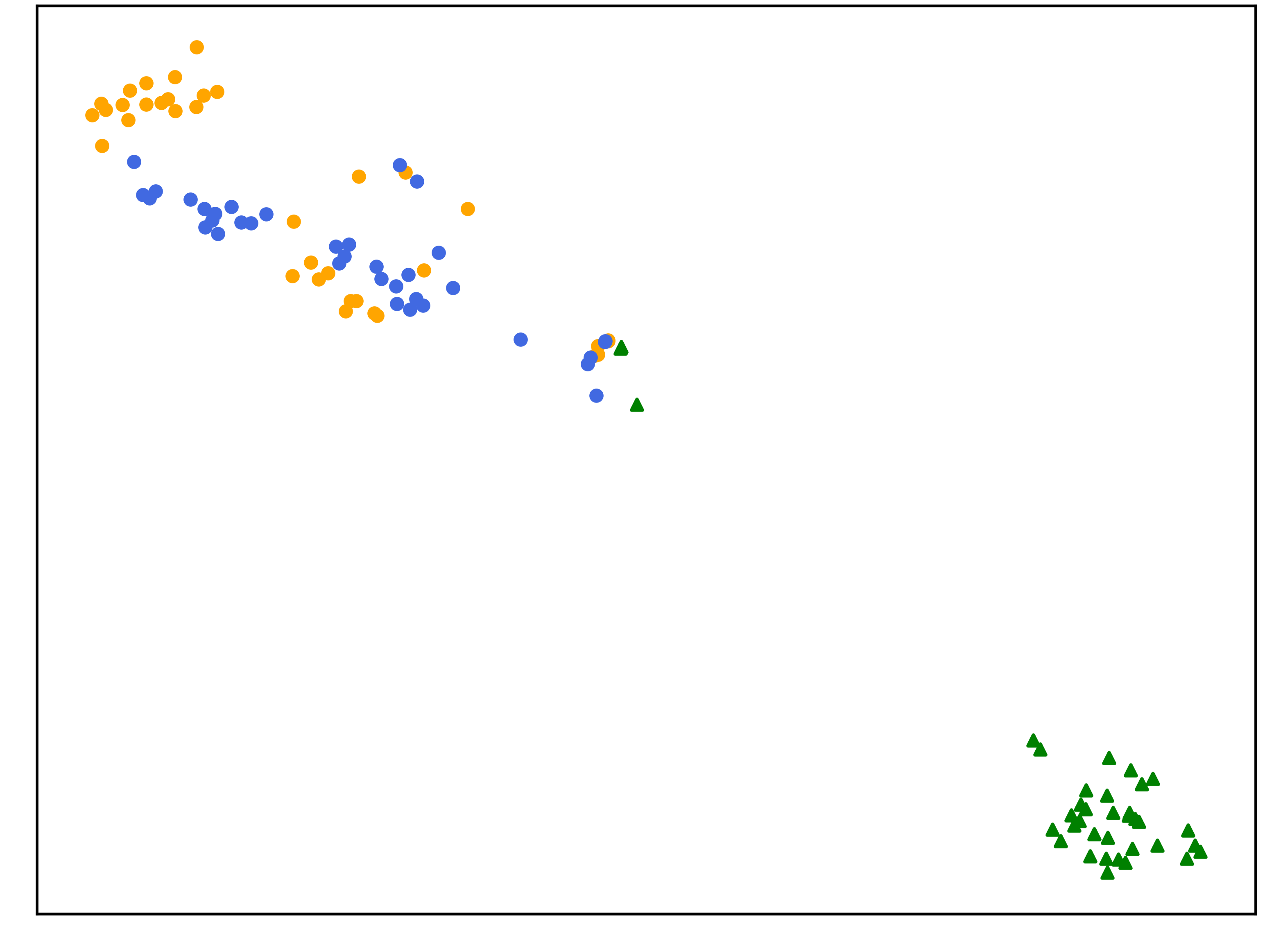}
        \centerline{ \small (a)w/o contrastive loss}
    \end{minipage}%
    \begin{minipage}[t]{0.5\linewidth}
        \centering
        \includegraphics[width=\textwidth]{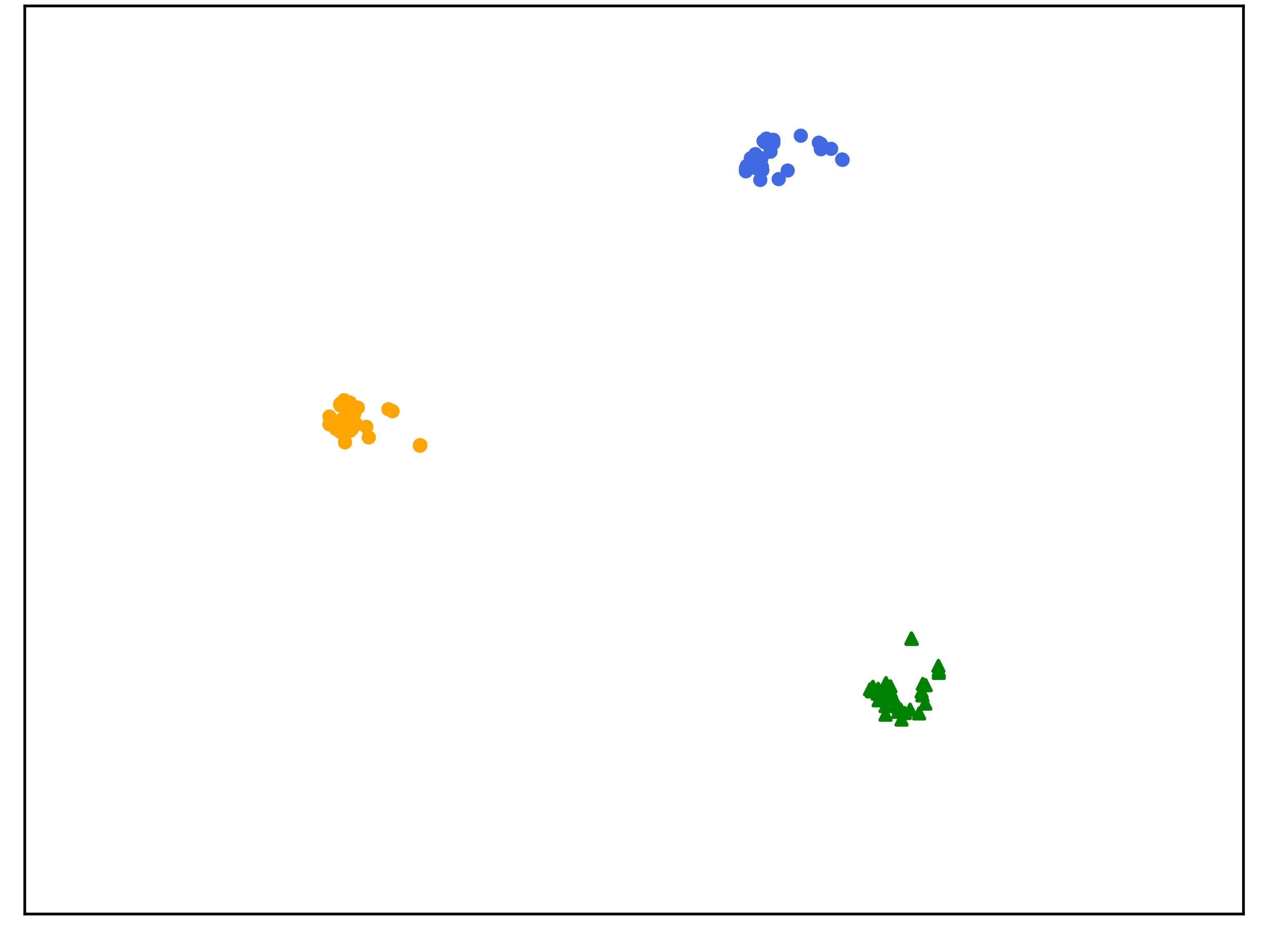}
        \centerline{ \small (b)w contrastive loss}
    \end{minipage}
    \caption{Visualization of the box queries with t-SNE~\cite{van2008visualizing}. The input video has three instances and two of them have the same category~(marked by yellow\&blue dots). It can be seen that with contrastive loss, the  box queries in the same category but belonging to different instances are pushed away.}
    \label{Fig. 6}
    \vspace{-3mm}
\end{figure}

\noindent \textbf{Contrastive learning. }\cref{Tab6}~ shows the effect of contrastive loss on frame-level box queries. We first apply the contrastive loss on the box queries generated by the last layer of the decoder. It can be seen that our proposed loss can boost the gain of $0.7\%$ AP. When training our model with auxiliary loss, we achieve the final $ \rm AP$ of $48.1\%$. To further illustrate the role played by contrastive loss, we visualize the box queries of the video named 00f88c4f0 in YTVIS-19 with t-SNE~\cite{van2008visualizing}, as shown in \cref{Fig. 6}. When contrastive loss is not adopted, although instances of different categories (marked by dots and triangles) are far away from each other, instances of the same category intersect in the embedding space, which leads to poor segmentation of two snuggly instances. Whereas, after adding contrastive loss, even the instances of the same category are separated from each other in the embedding space.

\noindent \textbf{Number of sampling points in ST-Enc.} It is also of interest to assess the effect of using how many temporal cues by adjusting the number of $K_{inter}$.  \cref{Tab7}~shows the results for this. We observe a large performance improvement when $K_{inter}$ changes from 1 to 3. When we further change $K_{inter}$ from 3 to 4, the gain of performance is not obvious. As a result, to achieve the balance between computation and precision, we set the value of $K_{inter}$ to 4 for all experiments. 

\begin{table}[h]\small
\centering
\renewcommand{\arraystretch}{1.1}
\begin{tabular}{ccc}
\hline
$K_{inter}$ & Params &$\rm AP$   \\ \hline
1           &  50.4   &  46.6 \\
2           &  51.0   &  47.5 \\
3           &   51.6  &  47.9  \\
4           &    52.2 & 48.1 \\ \hline
\end{tabular}
\caption{Ablation on the number of temporal sampling points.}
\label{Tab7}
\vspace{-3mm}
\end{table}

\section{Conclusion}
This paper presents TAFormer which aims to alleviate two problems in video instance segmentation: appearance deformation and tracking error caused by visually similar instances. For the appearance deformation, TAFormer aggregates spatio-temporal features by introducing a spatio-temporal join multi-scale deformable attention module in transformer encoder and a temporal self-attention module in transformer decoder. For the instance tracking error, TAFormer uses an instance-level contrastive loss to increase the discriminability of instance query. TAFormer improves state-of-the-art approaches on YouTube-VIS 2019/2021 benchmarks, which demonstrates the effectiveness of using temporal information in both transformer encoder and decoder. We hope our insights in TAFormer can inspire future transformer-based VIS methods.
{\small
\bibliographystyle{ieee_fullname}
\bibliography{egbib}
}

\end{document}